\pdfoutput=1

\documentclass[11pt]{article}

\usepackage{ACL2023}

\usepackage{times}
\usepackage{latexsym}

\usepackage[T1]{fontenc}

\usepackage[utf8]{inputenc}

\usepackage{microtype}
\usepackage{subcaption,graphicx,dblfloatfix}

\usepackage{inconsolata}

\usepackage{mathtools}
\usepackage{amssymb}
\usepackage{colortbl}
\usepackage{outline}
\usepackage{amsmath}
\usepackage{amsfonts}

\newcommand{\B}[1]{\textbf{#1}}
\newcommand{\U}[1]{\underline{#1}}

\definecolor{mypurple}{HTML}{9F00FF} 
\definecolor{myyellow}{HTML}{BA9500} 
\definecolor{myred}{HTML}{920500}
\definecolor{mygreen}{HTML}{3F7C0D}
\definecolor{myblue}{HTML}{3848D1}

\usepackage{bm}      
\usepackage{booktabs} 
\usepackage{multirow} 
\DeclareMathOperator*{\minimize}{min}

\title{Analysis of Utterance Embeddings and Clustering Methods Related to Intent Induction for Task-Oriented Dialogue}

\author{Jeiyoon Park$^{1,3}$, Yoonna Jang$^{1}$, Chanhee Lee$^{2}$, Heuiseok Lim$^{1}$ \\
  $^1$Department of Computer Science and Engineering, Korea University \\
  $^2$Naver Corporation \\
  $^3$LLSOLLU \\
  \{k4ke, morelychee, limhseok\}@korea.ac.kr, chanhee.lee@navercorp.com \\}

\begin{document}
\maketitle
\begin{abstract}
The focus of this work is to investigate unsupervised approaches to overcome quintessential challenges in designing task-oriented dialog schema: assigning intent labels to each dialog turn (\textit{intent clustering}) and generating a set of intents based on the intent clustering methods (\textit{intent induction}). We postulate there are two salient factors for automatic induction of intents: (1) clustering algorithm for intent labeling and (2) user utterance embedding space. We compare existing off-the-shelf clustering models and embeddings based on DSTC11 evaluation. Our extensive experiments demonstrate that the combined selection of utterance embedding and clustering method in the intent induction task should be carefully considered. We also present that pretrained MiniLM with Agglomerative clustering shows significant improvement in NMI, ARI, F1, accuracy and example coverage in intent induction tasks. The source codes are available at \url{https://github.com/Jeiyoon/dstc11-track2}.
\end{abstract}


\section{Introduction}

\textbf{Why Intent Induction?} We humans are generalists. During a conversation, we listen to the other person's utterance and naturally grasp which intent of the utterance is. With the skyrocketing demand for conversational AI, however, the more user utterances a dialogue system encounters, the more unknown intents it does. Predefining user intent is expensive and it is impossible to annotate all the user intents.

\begin{figure}[t]
    \centering
    \includegraphics[width=7.7cm]{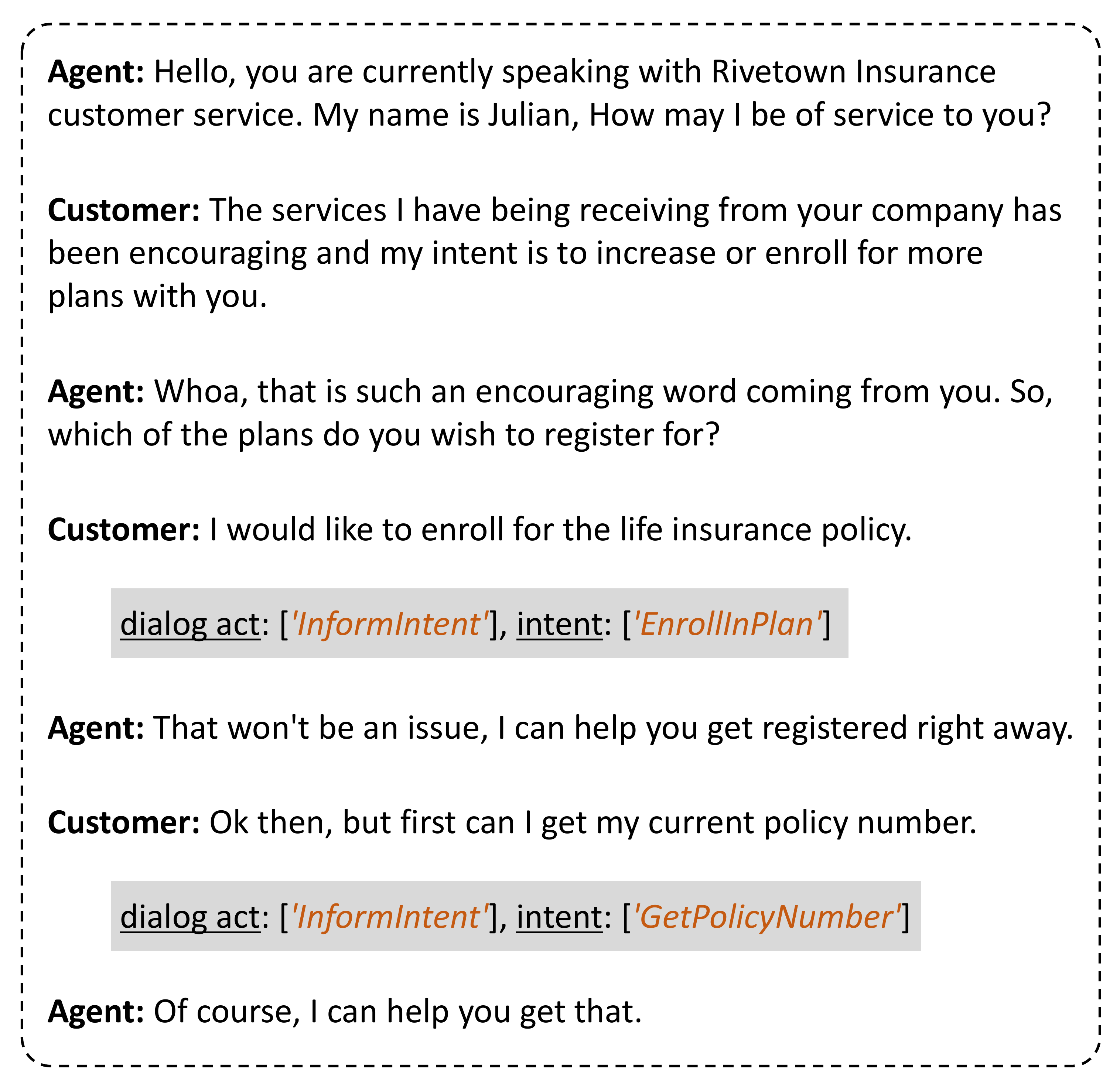}
    \caption{A sample segment of conversation transcript.}
    \label{fig:1_example}
\end{figure}
 
Since provided user utterances are unlabeled, intent induction \cite{haponchyk2018supervised,Perkins2019DialogII,chatterjeesengupta2020intent,10.1145/3442381.3450026,zhangetal2022new} focuses on discovering user intents from user utterances. However, previous studies did not conduct an in-depth analysis of the application of existing models to intent induction that might cause performance degradation problems \cite{10.1145/3442381.3450026}. 

\textbf{Our Approach.} Intuitively, for good intent induction, user utterances must be well represented in the embedding space, and good clustering algorithms must be employed to capture latent barycenters of user intent clusters to handle both predefined and unseen intents well. 

In this paper, We postulate there are two salient factors for automatic induction of intents: (1) clustering algorithm for intent labeling \cite{cheung2012sequence,hakkani2015clustering,padmasundari2018intent} and (2) user utterance embedding space \cite{NEURIPS2020_3f5ee243,NEURIPS2020_c3a690be,gaoetal2021simcse,chuangetal2022diffcse,nishikawaetal2022ease}. 

\begin{table}[t]
\centering
\small
\begin{tabular}{lccc}
\toprule
Dataset & Domain  & \#Intents & \#Utterances \\
\midrule
DSTC11$_{dev}$       & \textit{insurance} & 22   & 66,875 \\
DSTC11$_{test}$ & \textit{insurance} & 22   & 913 \\
\bottomrule
\end{tabular}
\caption{Statistics of development dataset.}
\label{tab:Dataset}
\end{table}

We analyze how the two key factors affect to user intent clustering and intent induction. Our extensive experiments with existing models demonstrate that pretrained MiniLM with Agglomerative clustering shows significant improvement in NMI, ARI, F1, accuracy and example coverage.

\section{Task 1: Intent Clustering}
\subsection{Task Description.} A set of conversation transcripts are given as Figure \ref{fig:1_example}, intent clustering model aims to (i) generate intent labels and (ii) align each utterance annotated with dialog act (i.e., \textit{"InformIntent"}). 

\textbf{Dataset.} We conduct experiments on DSTC11 development dataset. It consists of 948 human-to-human conversation transcript with speaker role, utterance, dialog act, and intent. Testset is composed of 913 customer utterances and corresponding user intents. The dataset statistics are summarized in Table \ref{tab:Dataset}.  

\textbf{Metrics.} We follow the same experimental metrics employed in the DSTC11 task proposal: Normalized mutual information (NMI), Adjusted rand index (ARI), Accuracy (ACC), Precision, Recall, F1 score, Intent example coverage, and the number of clusters (\#K).

NMI is used for measuring dependency between two different distributions:
\begin{align}
    \label{equation_1}
    NMI(X, Y) = \frac{\mathbb{I}(X;Y)}{\displaystyle{\minimize(\mathbb{H}(X), \mathbb{H}(Y))}}
\end{align}
, where $X = [X_1, ..., X_r]$ denote clustered labels, $Y = [Y_1, ..., Y_s]$ are reference labels, $\mathbb{I}$ stands for mutual information, and $\mathbb{H}$ is entropy. 

ARI is a measure for computing similarities between clustered results and reference labels:
\begin{align}
    \label{equation_2}
    ARI(X, Y) = \frac{\sum_{ij}\binom{n_{ij}}{2} - [AB]/\binom{n}{2}}{\frac{1}{2}[A + B] - [AB]/\binom{n}{2}}
\end{align}
, where $A = \sum_{i}\binom{a_i}{2}$, $B = \sum_{j}\binom{b_j}{2}$, $n_{ij} = |X_{i} \cap Y_{j}|$, $a_{i}$ is the number of clustered label $X_i$ and $b_{j}$ is the number of reference label $Y_j$.

Both Precision and Recall measure many-to-one alignments from clustered labels to reference labels. F1 score is a harmonic mean between precision and recall. After performing a many-to-one alignment, Intent example coverage is determined as percent of examples whose reference has a corresponding predicted result.

\subsection{Methods}
 
\textbf{Clustering Algorithm.} Intent clustering focuses on assigning dialog intents to each dialog. $K$-means clustering \cite{journalstitLloyd82}, for example, regards a set of conversation transcript $\mathcal{T}_{1}, ..., \mathcal{T}_{m}$. Each transcript $\mathcal{T}$ consists of turn-level dialog acts, speaker's role, whether it is Agent or Customer, and dialog utterances $X_{1}, ..., X_{n} \in \mathbb{R}^{p}$. $K$-means minimizes the following equation in the Euclidean embedding space:
\begin{align}
    \label{equation_3}
    \displaystyle{\minimize_{\beta_{1}, ...\beta_{K}\in \mathbb{R}^{d}} \displaystyle\sum_{i=1} ^{n}  \displaystyle{\minimize_{k \in [K] }} \mu_{ik}||X_i - \beta_k||_{2}^{2}}
\end{align}
, where $\{\beta_k\}_{k=1}^{K}$ denotes centroid of each intent cluster, $[K] = {1, 2, ..., K}$, $\mu_{ik}$ is alignment factor. Equation \ref{equation_3} can be expressed as:
\begin{align}
    \label{equation_4}
    \displaystyle{\minimize_{V_{1}, ...V_{K}} 
    \left\{ \displaystyle\sum_{k=1} ^{K} \displaystyle\sum_{i \in V_k} \mu_{ik}||X_i - \beta_k||_{2}^{2} : \displaystyle\bigsqcup_{k=1}^{K} V_k = [n] \right\}} 
\end{align}
, where $\sqcup$ stands for disjoint union, and $\{V_k \}_{k=1}^{K}$ is intent cluster, determined by Voronoi diagram in utterance embedding space.

Given initial intent centroids $\{\beta_{k}^{1}\}_{k=1}^K$, we assign each utterance embedding to its nearest centroid (a.k.a., Expectation step):
\begin{align}
    \label{equation_5}
    V_{k}^{t} = \left\{ i \in [n] : ||X_i - \beta_k||_{2}^{2} \leq ||X_i - \beta_j||_{2}^{2}, \right\}
\end{align}
, where $\forall j \in [K]$. Then, we update the location of centroids (a.k.a., Maximization step):
\begin{align}
    \label{equation_6}
    \beta_{k}^{t+1} = \frac{1}{|V_{k}^{t}|} \displaystyle\sum_{i \in V_{k}^{t}} X_i
\end{align}
We iterate equation \ref{equation_5} and equation \ref{equation_6} alternatively until equation \ref{equation_4} converges.

In this paper, we conduct experiments with $K$-means \cite{journalstitLloyd82}, BIRCH \cite{233324}, Agglomerative clustering \cite{steinbach00comparison}, Spectral clustering \cite{conficcvYuS03}, Bisecting $K$-means \cite{di2018bisecting}, and Variational optimal transportation (VOT) \cite{mi2018variational}.

\begin{table*}[ht!]
\definecolor{mygray-bg}{RGB}{211,211,211}
\centering
\resizebox{\textwidth}{!}{%
\begin{tabular}{l c c c c c c c c c c c c c}
\toprule
\B{Clustering Method} & \multicolumn{6}{c}{\B{$K$-means Clustering}}\\
\midrule
\B{Metric}              & \B{NMI}    & \B{ARI}    & \B{ACC}    & \B{Precision}  & \B{Recall}      & \B{F1} & \B{Example Coverage} & \B{\#K}\\
\midrule
EASE$_{ROBERTA}^{m}$ &  33.4  &  14.2  &  28.0  &  28.0  &  \textbf{69.0}  &  39.9  &  43.9  &  5 \\
EASE$_{BERT}$ &  36.1  &  18.1  &  35.3  &  36.8  &  58.9  &  45.3  &  53.5  &  8 \\
EASE$_{BERT}^{m}$ &  38.4  &  10.9  &  23.6  &  42.7  &  24.2  &  30.9  &  87.6  &  44 \\
EASE$_{ROBERTA}$ &  45.8  &  25.4  &  40.0  &  43.7  &  60.7  &  50.9  &  65.6  &  12 \\
\midrule
DiffCSE$_{BERT}^{trans}$ &  43.8  &  15.5  &  29.4  &  49.5  &  30.6  &  37.8  &  90.1  &  40 \\
DiffCSE$_{BERT}^{sts}$ &  46.6  &  16.9  &  29.9  &  50.4  &  30.8  &  38.2  &  89.9  &  43 \\
DiffCSE$_{ROBERTA}^{sts}$ &  53.9  &  29.0  &  45.1  &  57.2  &  46.8  &  51.5  &  91.1  &  30   \\
DiffCSE$_{ROBERTA}^{trans}$ &  55.0  &  23.1  &  35.7  &  60.6  &  35.7  &  44.9  &  96.7  &  50   \\
\midrule
SimCSE$_{BERT}^{u}$ &  31.7  &  13.3  &  27.9  &  27.9  &  \U{65.0}  &  39.0  &  43.9  &  5 \\
SimCSE$_{BERT+}^{u}$ &  47.0  &  25.7  &  38.4  &  46.9  &  46.9  &  46.9  &  76.8  &  19  \\
SimCSE$_{ROBERTA}^{u}$ &  51.2  &  29.0  &  44.4  &  49.7  &  61.6  &  55.0  &  70.5  &  14   \\
SimCSE$_{BERT+}$ &  53.0  &  27.8  &  39.8  &  55.0  &  42.2  &  47.8  &  91.0  &  31 \\
SimCSE$_{BERT}$ &  53.1  &  24.4  &  39.0  &  60.5  &  39.5  &  47.8  &  96.3  &  44   \\
SimCSE$_{ROBERTA+}^{u}$ &  53.2  &  25.9  &  42.2  &  56.8  &  43.8  &  49.5  &  91.7  &  32   \\
SimCSE$_{ROBERTA}$ & 56.6 & 28.8     & 42.7     & 60.8     & 43.3     & 50.6 & 91.3     & 36  \\
SimCSE$_{ROBERTA+}$    & 56.8     & 28.9     & 41.3     & 62.5    & 41.6     & 49.9   & 98.4     & 42\\
\midrule
Glove$^{avg}$ &  30.5  &  7.0  &  20.6  &  34.6  &  22.2  &  27.0  &  92.2  &  50 \\
MPNet       & 59.3     & 32.3     & 46.1     & 66.0     & 47.1     & 54.9      & 96.5      & 42 \\
MiniLM$_{L6}$ &  59.3  &  35.7  &  52.6  &  62.2  &  54.9  &  58.4  &  92.4  &  28 \\
\cellcolor{mygray-bg}MiniLM$_{MULTIQA}$ &  \cellcolor{mygray-bg}\U{61.7}  &  \cellcolor{mygray-bg}\U{38.2}  &  \cellcolor{mygray-bg}\textbf{55.1}  &  \cellcolor{mygray-bg}\U{66.6}  &  \cellcolor{mygray-bg}55.4  &  \cellcolor{mygray-bg}\U{60.5}  &  \cellcolor{mygray-bg}\U{98.8}  &  \cellcolor{mygray-bg}30 \\
\cellcolor{mygray-bg}MiniLM$_{L12}$      & \cellcolor{mygray-bg}\textbf{63.1}     & \cellcolor{mygray-bg}\textbf{38.9}     & \cellcolor{mygray-bg}\U{54.9}     & \cellcolor{mygray-bg}\textbf{68.0}     & \cellcolor{mygray-bg}54.9     & \cellcolor{mygray-bg}\textbf{60.8}   & \cellcolor{mygray-bg}\textbf{100.0}     & \cellcolor{mygray-bg}31 \\
\bottomrule
\end{tabular}
}
\caption{Clustering results on DSTC11 dataset. We employ $K$-means clustering algorithm to all utterance embeddings. $m$ denotes multilingual model, $u$ stands for unsupervised model, and $+$ means large model.}
\label{table:cluster_results_emb}
 \vspace{-.1in}
\end{table*}

\textbf{Embeddings.} User utterance should be represented in the embedding space which is able to capture universal and rich semantic information. MiniLM \cite{NEURIPS2020_3f5ee243} is an effective task-agnostic distillation to compress transformer-based language models. MPNet \cite{NEURIPS2020_c3a690be} proposes permuted language model for dependency among predicted tokens and makes the model to see a full sentence and auxiliary position information. SimCSE \cite{gaoetal2021simcse} leverages a simple contrastive learning framework. DiffCSE \cite{chuangetal2022diffcse} learns the difference between the original sentence and a stochastically masked sentence. EASE \cite{nishikawaetal2022ease} exploits sentence embedding via contrastive learning between the original sentence and its related entities.      

\subsection{Result Analysis}
To analyze the effects of both embedding and clustering algorithm, We first heuristically fix the clustering algorithm and find the most meaningful embedding. Then, we opt for the most suitable clustering method based on the embedding.

\begin{table}[t]
\centering
\small
\begin{tabular}{lcccc}
\toprule
Method & DiffCSE  & SimCSE & MPNet & MiniLM \\
\midrule
\# Param       & 250M & 125M   & 110M  & 21.3M\\
\bottomrule
\end{tabular}
\caption{The number of parameters. Both DiffCSE and SimCSE denote RoBERTa\textsubscript{base} model.}
\label{tab:Parameter}
\end{table}

\textbf{Analysis of Embeddings.} We show the experimental results for analyzing the effect of embedding space in Table \ref{table:cluster_results_emb}. We observe that EASE records poor performance in all metrics, followed by averaged Glove embedding which shows the worst result, though EASE is a large-scale model. Note that both entity-aware contrastive learning and multilingual setting cause performance degradation in dialog intent clustering task. 

\begin{figure*}[!ht]
 \centering
    \includegraphics[scale=0.155]{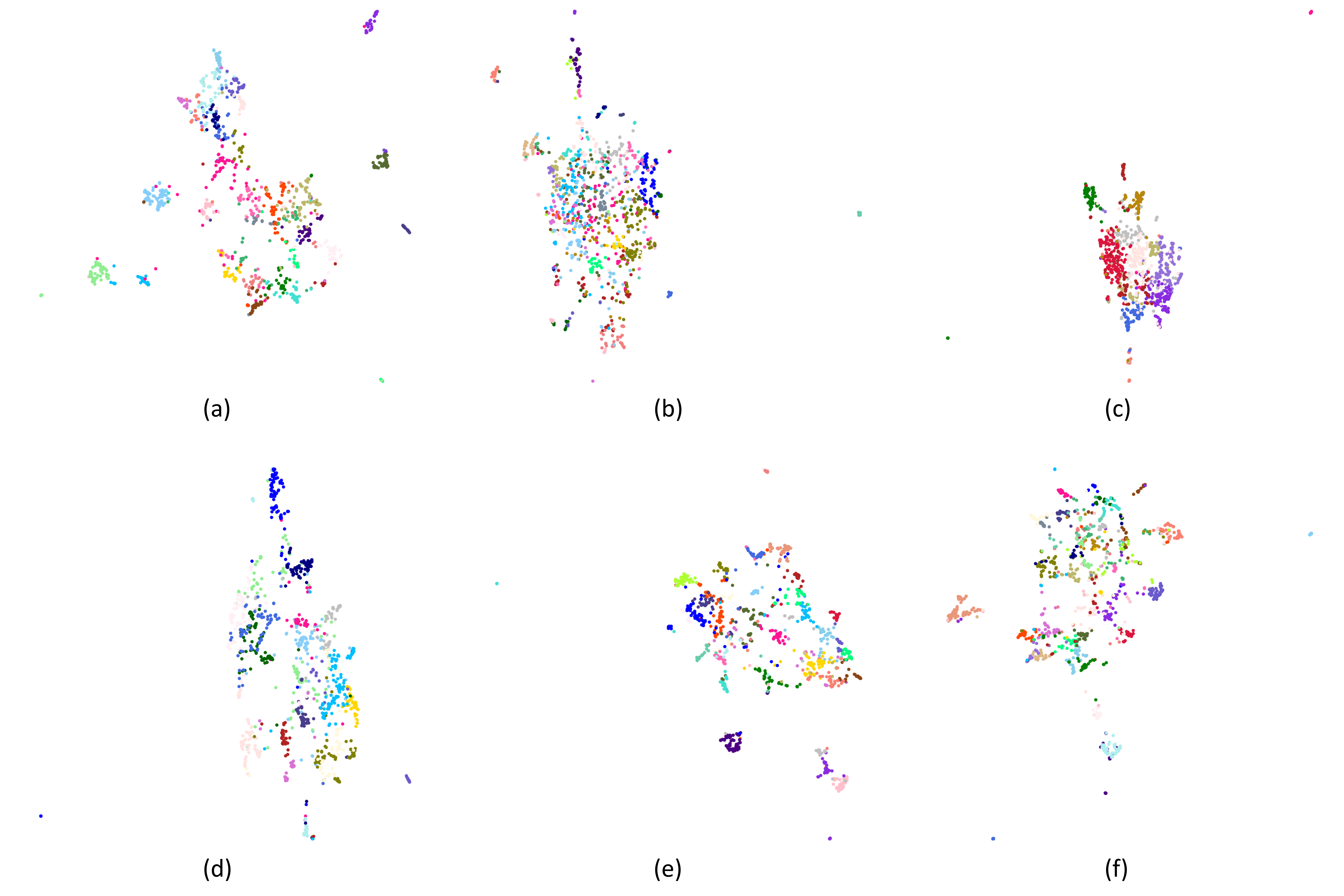}
 \captionof{figure}{\label{fig:2_scatter} UMAP visualization of intent clustering results with different embeddings based on DSTC11 development dataset: (a) MiniLM, (b) Glove, (c) Ease-roberta, (d) DiffCSE-roberta, (e) SimCSE-roberta, and (f) MPNet. We apply $K$-means clustering algorithm to (a)-(e) experiments.}
 \vspace{-.1in}
\end{figure*}

\begin{table*}[ht!]
\definecolor{mygray-bg}{RGB}{211,211,211}
\centering
\resizebox{\textwidth}{!}{%
\begin{tabular}{l c c c c c c c c c c c c c}
\toprule
\B{Utterance Embedding} & \multicolumn{6}{c}{\B{MiniLM}\textsubscript{L12}}\\
\midrule
\B{Metric}              & \B{NMI}    & \B{ARI}    & \B{ACC}    & \B{Precision}  & \B{Recall}      & \B{F1} & \B{Example Coverage} & \B{\#K}\\
\midrule
Bisect $K$-means &  32.2  &  13.9  &  23.8  &  26.0  &  \textbf{67.4}  &  37.5  &  37.0  &  5 \\
VOT &  53.8  &  31.8  &  48.5  &  54.1  &  53.5  &  53.8  &  84.8  &  20 \\
Spectral       &  57.6  &  \U{34.9}  &  51.3  &  58.8  &  \U{56.1}  &  \U{57.4}  &  82.9  &  24 \\
Agglomerative &  57.9  &  34.2  &  \U{51.8}  &  58.0  &  55.4  &  56.7  &  \U{88.6}    &  23 \\
BIRCH &  \U{59.9}  &  32.9  &  46.3  &  \U{64.6}  &  47.0  &  54.4  &  \textbf{100.0}  &  47 \\
\cellcolor{mygray-bg}$K$-means      & \cellcolor{mygray-bg}\textbf{63.1}     & \cellcolor{mygray-bg}\textbf{38.9}     & \cellcolor{mygray-bg}\textbf{54.9}     & \cellcolor{mygray-bg}\textbf{68.0}     & \cellcolor{mygray-bg}54.9     & \cellcolor{mygray-bg}\textbf{60.8}   & \cellcolor{mygray-bg}\textbf{100.0}     & \cellcolor{mygray-bg}31 \\
\bottomrule
\end{tabular}
}
\caption{Clustering results on DSTC11 dataset. We apply MiniLM\textsubscript{L12} utterance embedding to all clustering methods.}
\label{table:cluster_results_cltd}
 \vspace{-.1in}
\end{table*}

The results for DiffCSE also show that unsupervised contrastive learning between original utterance and edited utterance exacerbates model performance in both STS and Trans tasks. Despite the model size being twice smaller than DiffCSE as shown in Table \ref{tab:Parameter}, SimCSE gets comparable scores to MPNet, with a similar model size. Note that MiniLM achieves remarkable performances in both $L12$ setting\footnote{\url{https://huggingface.co/sentence-transformers/all-MiniLM-L12-v2}} and $MULTIQA$ setting\footnote{\url{https://huggingface.co/sentence-transformers/multi-qa-MiniLM-L6-cos-v1}} in all metrics. These results demonstrate that (i) performance increases as the number of parameter decreases, which means excessively large embedding model leads to performance degradation, and (ii) the use of a student network which is trained by the teacher's self-attention distributions and guiding layer improves intent clustering performance.

\textbf{Visualization.} Figure \ref{fig:2_scatter} gives UMAP \cite{2018arXivUMAP} visualization of clustering results. Note that we employ UMAP, instead of t-SNE \cite{Maaten2008VisualizingDU}, because UMAP preserves more global structure than t-SNE. We observe that (b) - (d) vertically degenerated into each embedding space. We also find that (a) presents the most well-clustered result and covers all intents (i.e., Example Coverage is 100.0) while (e) and (f) embeddings suffer from relatively ill-clustered result and outliers.  

\textbf{Analysis of Clustering Methods.} In Table \ref{table:cluster_results_cltd}, we observe that $K$-means method outperforms the other models, followed by BIRCH. Note that VOT records poor performance which means optimal transportation with the variational principle deteriorates the result in intent clustering task. Bisect $K$-means clustering algorithm calculates the point density and average density of all points to initialize cluster barycenters. However, $K$-means with $K$-means++ \cite{10.5555/1283383.1283494} initializer shows much better performance, contributing to the overall cluster inertia.

\begin{figure*}[!ht]
 \centering
    \includegraphics[scale=0.155]{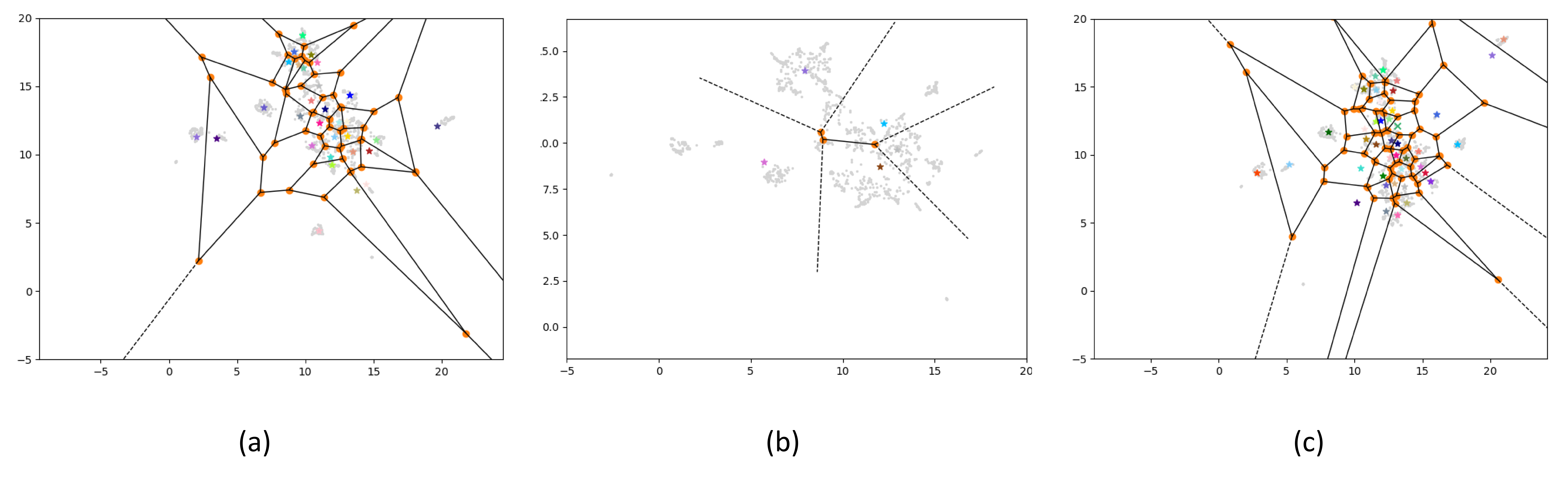}
 \captionof{figure}{\label{fig:3_scatter} Voronoi diagram visualization of intent clustering results with different clustering algorithms based on DSTC11 development dataset: (a) $K$-means, (b) Bisect $K$-means, and (c) BIRCH. We apply MiniLM\textsubscript{L12} user utterance embedding to (a)-(c) experiments. * stands for barycenter of each cluster and x represents outlier barycenter with fewer than five utterances.}
 \vspace{-.1in}
\end{figure*}

\begin{table*}[ht!]
\definecolor{mygray-bg}{RGB}{211,211,211}
\centering
\resizebox{\textwidth}{!}{%
\begin{tabular}{l c c c c c c c c c c c c c}
\toprule
\B{Clustering Method} & \multicolumn{6}{c}{\B{$K$-means Clustering}}\\
\midrule
\B{Metric}              & \B{NMI}    & \B{ARI}    & \B{ACC}    & \B{Precision}  & \B{Recall}      & \B{F1} & \B{Example Coverage} & \B{\#K}\\
\midrule
EASE$_{BERT}^{m}$ &  27.0  &  12.3  &  23.9  &  24.1  &  72.9  &  36.2  &  26.4  &  5 \\
EASE$_{BERT}$ &  53.1  &  25.6  &  42.8  &  50.2  &  57.4  &  53.5  &  80.0  &  26 \\
EASE$_{ROBERTA}^{m}$ &  59.6  &  41.8  &  53.7  &  57.8  &  68.8  &  62.8  &  86.0  &  34 \\
EASE$_{ROBERTA}$ &  60.5  &  50.7  &  52.7  &  55.8  &  70.0  &  62.1  &  83.1  &  28 \\
\midrule
DiffCSE$_{BERT}^{trans}$ &  21.0  &  11.5  &  23.3  &  23.7  &  \U{77.5}  &  36.3  &  32.5  &  6 \\
DiffCSE$_{BERT}^{sts}$ &  49.6  &  22.3  &  40.2  &  46.1  &  58.4  &  51.5  &  93.2  &  31 \\
DiffCSE$_{ROBERTA}^{sts}$ &  65.4  &  42.3  &  53.7  &  59.4  &  65.9  &  62.5  &  89.8  &  31   \\
DiffCSE$_{ROBERTA}^{trans}$ &  65.4  &  53.3  &  57.6  &  63.1  &  71.1  &  66.8  &  \U{90.0}  &  28   \\
\midrule
SimCSE$_{BERT+}^{u}$ &  33.1  &  19.1  &  28.0  &  28.0  &  \U{77.5}  &  41.2  &  33.4  &  5  \\
SimCSE$_{BERT}^{u}$ &  58.3  &  29.6  &  46.4  &  58.5  &  59.1  &  58.8  &  83.1  &  32 \\
SimCSE$_{BERT+}$ &  59.4  &  29.1  &  47.0  &  54.2  &  60.5  &  57.2  &  86.2  &  30 \\
SimCSE$_{ROBERTA}$ &  61.4  &  34.3  &  47.1  &  54.3  &  62.0  &  57.9  &  80.3  &  29  \\
SimCSE$_{BERT}$ &  62.7  &  37.3  &  52.6  &  63.4  &  59.8  &  61.6  &  89.6  &  34   \\
SimCSE$_{ROBERTA+}^{u}$ &  67.8  &  51.1  &  57.7  &  64.2  &  70.2  &  67.1  &  86.4  &  32   \\
SimCSE$_{ROBERTA}^{u}$ &  68.6  &  48.0  &  54.7  &  65.6  &  71.2  &  68.3  &  86.1  &  31   \\
SimCSE$_{ROBERTA+}$    &  69.2  &  38.7  &  52.1  &  62.4  &  70.0  &  66.0  &  \textbf{93.8}  &  32\\
\midrule
Glove$^{avg}$ &  35.0  &  18.8  &  29.1  &  36.3  &  51.2  &  42.4  &  77.0  &  29 \\
MPNet       &  72.8  &  41.6  &  60.2  &  65.6  &  76.3  &  70.6  &  86.5  &  26 \\
MiniLM$_{L12}$ &   73.2  &  47.1  &  57.1  &  \U{66.0}  &  72.2  &  69.0  &  83.1  &  25 \\
MiniLM$_{L6}$ &  \U{74.7}  &  \U{52.8}  &  \U{61.2}  &  \textbf{70.5}  &  74.9  &  \U{72.7}  &  83.4  &  25 \\
\cellcolor{mygray-bg}MiniLM$_{MULTIQA}$ &  \cellcolor{mygray-bg}\textbf{77.4}  &  \cellcolor{mygray-bg}\textbf{54.5}  &  \cellcolor{mygray-bg}\textbf{63.2}  &  \cellcolor{mygray-bg}\textbf{70.5}  &  \cellcolor{mygray-bg}\textbf{79.1}  &  \cellcolor{mygray-bg}\textbf{74.6}  &  \cellcolor{mygray-bg}80.0  &  \cellcolor{mygray-bg}24 \\
\bottomrule
\end{tabular}
}
\caption{Intent induction results on DSTC11 dataset. We employ $K$-means clustering algorithm to all utterance embeddings. $m$ denotes multilingual model, $u$ stands for unsupervised model, and $+$ means large model.}
\label{table:task2_cluster_results}
 \vspace{-.1in}
\end{table*}

\textbf{Voronoi Diagram.} Figure \ref{fig:3_scatter} gives UMAP visualization of clustering results with Voronoi diagram. The number of barycenters of Bisect $K$-means is five which means this ill-clustering model is not able to cover all latent user intents. Though BIRCH records compatible results to $K$-means, the number of predicted barycenters is excessively large compared to the number of reference $K$. It causes outlier barycenter problem that barycenter contains a few utterances and disrupts universal clustering representation. It also demonstrates that hierarchical process of BIRCH including removing outliers and cluster refining does not have a substantial impact.

\begin{figure*}[!ht]
 \centering
    \includegraphics[scale=0.15]{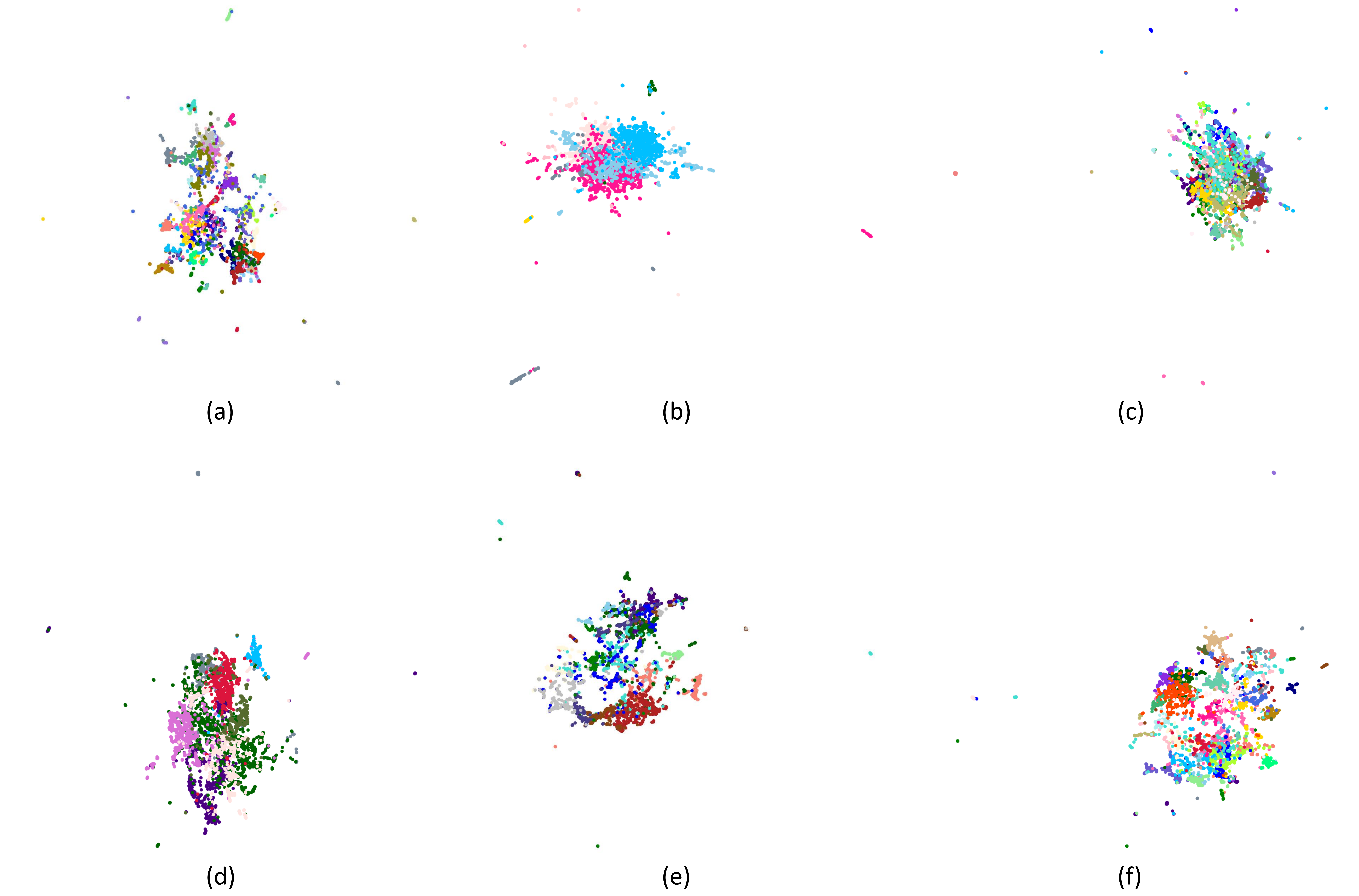}
 \captionof{figure}{\label{fig:4_scatter_task2} UMAP visualization of intent induction results with different embeddings based on DSTC11 development dataset: (a) MiniLM, (b) Glove, (c) Ease-roberta, (d) DiffCSE-roberta, (e) SimCSE-roberta, and (f) MPNet. We apply Agglomerative clustering algorithm to (a)-(e) experiments.}
 \vspace{-.1in}
\end{figure*}

\begin{table*}[ht!]
\definecolor{mygray-bg}{RGB}{211,211,211}
\centering
\resizebox{\textwidth}{!}{%
\begin{tabular}{l c c c c c c c c c c c c c}
\toprule
\B{Utterance Embedding} & \multicolumn{6}{c}{\B{MiniLM}\textsubscript{MULTIQA}}\\
\midrule
\B{Metric}              & \B{NMI}    & \B{ARI}    & \B{ACC}    & \B{Precision}  & \B{Recall}      & \B{F1} & \B{Example Coverage} & \B{\#K}\\
\midrule
Bisect $K$-means &  72.3  &  55.7  &  57.4  &  63.0  &  \U{81.2}  &  70.9  &  76.8  &  27 \\
VOT &  75.3  &  50.6  &  60.0  &  71.9  &  72.3  &  72.1  &  \textbf{93.4}  &  33 \\
Spectral       &  75.0  &  51.7  &  59.0  &  67.7  &  75.9  &  71.6  &  86.3  &  23 \\
$K$-means &  77.4  &  54.5  &  63.2  &  70.5  &  79.1  &  74.6  &  80.0  &  24\\
\cellcolor{mygray-bg}BIRCH &  \cellcolor{mygray-bg}\U{79.5}  &  \cellcolor{mygray-bg}\U{62.8}  &  \cellcolor{mygray-bg}\textbf{68.1}  &  \cellcolor{mygray-bg}\U{71.9}  &  \cellcolor{mygray-bg}\textbf{84.6}  &  \cellcolor{mygray-bg}\U{77.7}  &  \cellcolor{mygray-bg}\U{89.9}  &  \cellcolor{mygray-bg}24 \\
\cellcolor{mygray-bg}Agglomerative &  \cellcolor{mygray-bg}\textbf{81.0}  &  \cellcolor{mygray-bg}\textbf{64.2}  &  \cellcolor{mygray-bg}\U{66.5}  &  \cellcolor{mygray-bg}\textbf{75.5}  &  \cellcolor{mygray-bg}80.6  &  \cellcolor{mygray-bg}\textbf{78.0}  &  \cellcolor{mygray-bg}86.2  &  \cellcolor{mygray-bg}25 \\
\bottomrule
\end{tabular}
}
\caption{Intent induction results on DSTC11 dataset. We apply MiniLM\textsubscript{MULTIQA} utterance embedding to all clustering methods.}
\label{table:cluster_results_task2}
 \vspace{-.1in}
\end{table*}

\section{Task 2: Intent Induction}

\subsection{Task Description}

\begin{figure*}[!ht]
 \centering
    \includegraphics[scale=0.155]{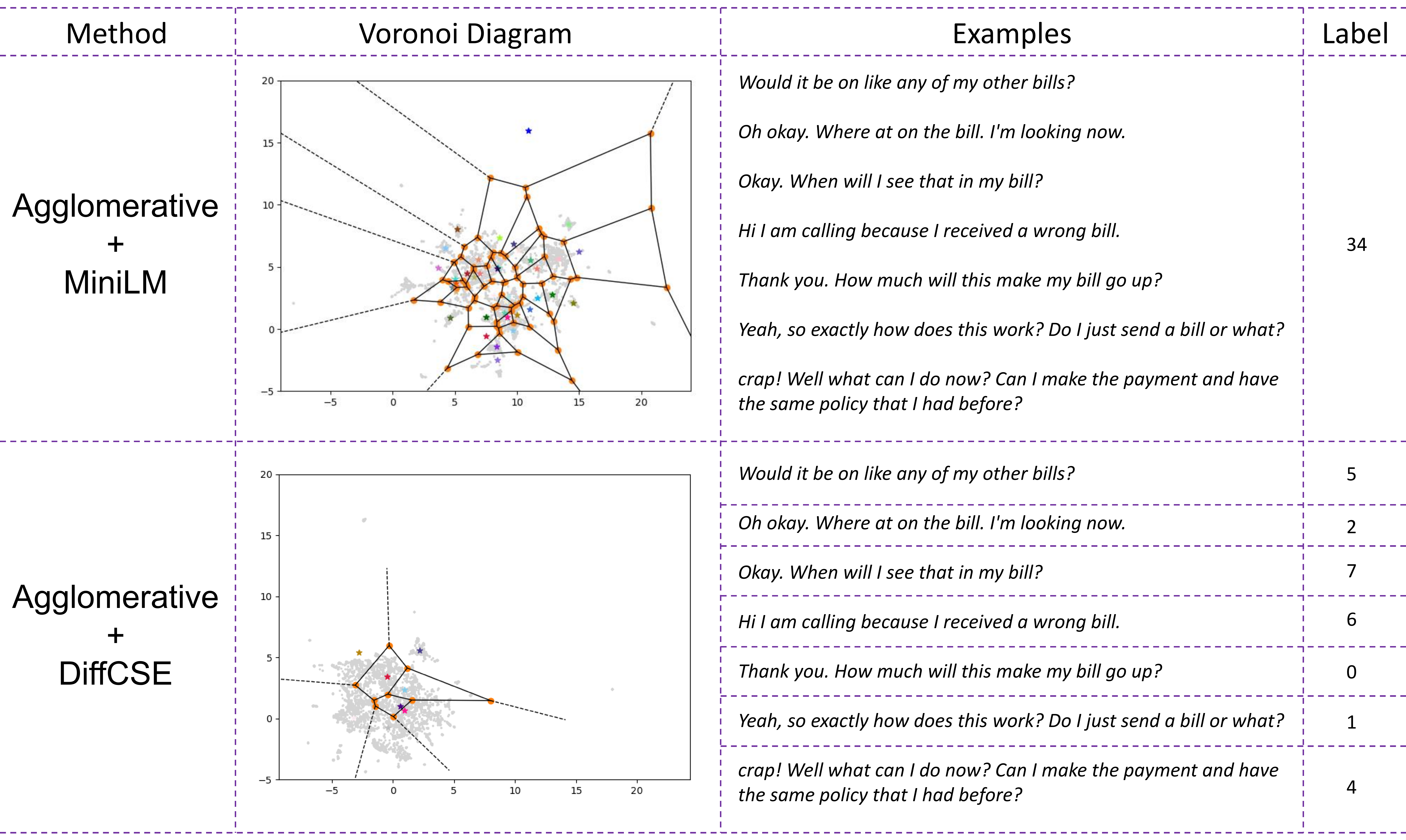}
 \captionof{figure}{\label{fig:5_qualitative} Qualitative result on DSTC11 dataset comparing MiniLM-based and DiffCSE-based models. We fix the clustering model as Agglomerative clustering in this experiment.}
 \vspace{-.1in}
\end{figure*}

Unlike Task 1, intent induction takes different transcripts which annotate only the speaker's role and intent label. The goal of intent induction is to create a set of intents and match them to each utterance without access to the ground-truth dialog acts. In this paper, we use DSTC11 dataset as shown in Table \ref{tab:Dataset}, exploit the same clustering algorithms and utterance embeddings, and evaluate methods using NMI, ARI, Accuracy, F1, and Example coverage. We employ the provided automatic dialog act predictions. 

\subsection{Result Analysis}

\textbf{Analysis of Embeddings.} Table \ref{table:task2_cluster_results} demonstrates that MiniLM-based utterance embedding can improve induction performance for user intents. Unlike Task 1, MiniLM\textsubscript{MULTIQA} outperforms the other methods, followed by MiniLM\textsubscript{L6}\footnote{\url{https://huggingface.co/sentence-transformers/all-MiniLM-L6-v2}}. It represents that MiniLM with 6 layers and 384 hidden size can learn universal utterance representation and capture meaningful information. Though the performance tends to increase as the size of each embedding model increases, we observe that larger models do not always perform better. Note that DiffCSE$^{trans}_{BERT}$ and EASE$^{m}_{BERT}$ record high recall because the number of clusters $K$ is disastrously low. We observe that  it dampens overall model performance. As shown in both Task 1 and Task 2 results, (i) entity-aware contrastive learning, (ii) multilingual setting, (iii) contrastive learning between the original utterance and edited utterance are ultimately not the optimal choice to improve intent induction performance.     

\textbf{Visualization.} We present UMAP visualization of intent induction results with different utterance embeddings (Figure \ref{fig:4_scatter_task2}). It corroborates that MiniLM-based intent induction can provide wider and well-separated results preserving its meaningful embedding space.   

\textbf{Analysis of Clustering Methods.} In Table \ref{table:cluster_results_task2}, we show the results of intent induction comparing models that clustering algorithms. Unlike the results in Task 1, Agglomerative clustering and BIRCH outperform the other baselines. Agglomerative clustering starts with the utterances as individual clusters and merges them if they have similar intents. The experiment shows that the approach exploiting MiniLM\textsubscript{MULTIQA} and Agglomerative clustering achieves state-of-the-art results among the other methods in the intent induction task.

\subsection{Qualitative Results}

Figure \ref{fig:5_qualitative} shows the qualitative result for intent induction on DSTC11 dataset. We generate user intent labels using two different models. We observe that Agglomerative clustering on MiniLM utterance embedding can classify bill-related utterances well. However, Agglomerative clustering on DiffCSE utterance embedding is not able to discern user intent well and amalgamate completely different user utterances into a single intent cluster. Note that all clusters have bill-related user utterances which means it is an ill-clustered result. 

We conclude that user utterance embedding is one of the most important factors affecting performance, and we should add a huge caveat that selection of utterance embedding in intent induction task should be very careful. Further analyzing how to leverage an embedding model is an interesting direction for future work.

Our extensive experiments demonstrate that the combined selection of utterance embedding and clustering method in the intent induction task should be carefully considered.

\begin{figure*}[!ht]
 \centering
    \includegraphics[scale=0.155]{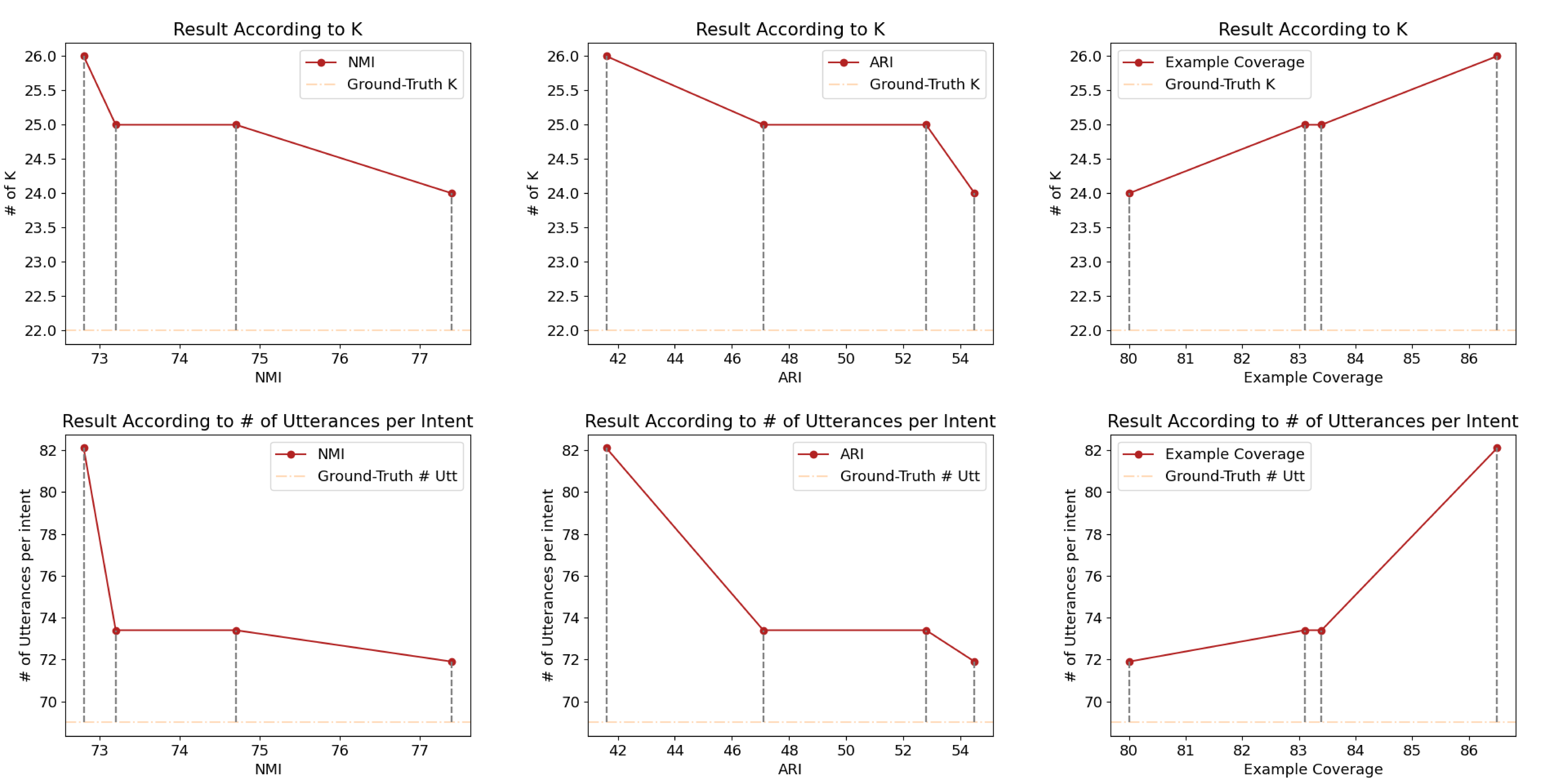}
 \captionof{figure}{\label{fig:6_metric} Quantitative results on DSTC11 dataset related to NMI, ARI, F1 score, and Example coverage. We analyze using four results with the best performance introduced in Table \ref{table:task2_cluster_results}.}
 \vspace{-.1in}
\end{figure*}

\subsection{Quantitative Results}

We conduct quantitative experiments to analyze (i) the correlation between the number of inducted clusters and performance, and (ii) the correlation between the number of utterances per intent and performance (Figure \ref{fig:6_metric}). Note that the maximum number of sample utterances aligned to intent is limited to 50. First, We find that NMI and ARI increase as the number of inducted clusters approaches the reference K. We also observe that Example coverage decreases as the number of inducted clusters go toward the reference K. We demonstrate that there is a trade-off relationship between NMI, ARI and Example coverage. Second, the correlation between the number of utterances per intent and performance shows the same pattern as the correlation between the number of inducted clusters and performance. 

\section{Conclusion}

The conclusion of this paper is threefold:
\begin{itemize}
  \setlength\itemsep{0em}
  \item We empirically demonstrate that the combined selection of utterance embedding and clustering method in the intent induction task should be carefully considered.
  \item We also present that pretrained MiniLM with Agglomerative clustering shows significant improvement in NMI, ARI, F1, accuracy and example coverage in intent induction tasks.
  \item We find that there is a trade-off relationship between NMI, ARI and Example coverage.
\end{itemize}

\section{Discussion and Future Work} 

In clustering, broadly, there are three categories of methods: (i) Barycenter-based formulation, (ii) Density-based formulation, and (iii) Distance-based formulation. In this paper, we mainly dealt with barycenter-based methods. Indeed, $K$-means clustering method, for instance, theoretically produce the same result as barycenter-based formulation and distacne-based formulation in Euclidean embedding space:
\begin{align}
    \label{equation_7}
    \displaystyle{\displaystyle\sum_{i,j=1} ^{n} ||X_i - X_j||_{2}^{2} = 2n\displaystyle\sum_{i=1} ^{n} ||X_i - \beta||_{2}^{2}} 
\end{align}
, where $\beta = \frac{1}{n}\sum_{i=1}^{n}X_i$. Besides, barycenter-based $K$-means in Euclidean space can circumvent the problem frequently caused by the location of barycenter in different measure space (e.g., Wasserstein space \cite{zhuang2022wasserstein}):  
\begin{equation}
\begin{aligned}
    \label{equation_8}
    \sum_{i=1}^{n} ||X - X_i||_{2}^{2} & = n||X - \beta||_{2}^{2} + \sum_{i=1}^{n} ||X - X_i||_{2}^{2} \\
    &\geqslant n||X - \beta||_{2}^{2}
\end{aligned}
\end{equation}
However, $K$-means clustering in Euclidean space often loses salient geometric information of dataset, due to its formulation.   
In the same vein, though we didn't add the experimental result of Density-based formulation to the table, the performance was disastrous. DBSCAN \cite{10.5555/3001460.3001507}, for example, recorded $7.7$ NMI, $0.0$ ARI, $16.1$ Accuracy, $27.6$ F1 score, and $41.0$ Example coverage in intent clustering task. Therefore, both clustering in different measure spaces and clustering using embedding density should be investigated. Further analyzing how to leverage an embedding model is also an interesting direction for future work.

\bibliography{anthology,custom}
\bibliographystyle{acl_natbib}

\section*{Acknowledgements}
This work was supported by Institute of Information \& communications Technology Planning \& Evaluation(IITP) grant funded by the Korea government(MSIT) (No. 2020-0-00368, A Neural-Symbolic Model for Knowledge Acquisition and Inference Techniques). This research was supported by the MSIT(Ministry of Science and ICT), Korea, under the ITRC(Information Technology Research Center) support program(IITP-2022-2018-0-01405) supervised by the IITP(Institute for Information \& Communications Technology Planning \& Evaluation).




\end{document}